\documentclass[letterpaper, 10 pt, conference]{ieeeconf}  
\usepackage[utf8]{inputenc} 
\usepackage[T1]{fontenc}    
\usepackage{hyperref}       
\usepackage{url}            
\usepackage{booktabs}       
\usepackage{amsfonts}       
\usepackage{nicefrac}       
\usepackage{microtype}      
\usepackage{xcolor}         
\usepackage{amsmath}
\usepackage{graphicx}
\usepackage{algorithm}
\usepackage{algpseudocode} 
\usepackage[font=small,labelfont=bf,skip = 2pt]{caption}
\usepackage{wrapfig}
\usepackage{multirow}

\usepackage{array} 

\IEEEoverridecommandlockouts                              

\title{\LARGE \bf Driving the Style: Velocity-Conditioned Whole-Body Control \\ from Motion Demonstrations}

\author{Murad Dawood$^{1}$ \and Oezge Drama Berkenkamp$^{1}$ \and Antonio Gonzales Marin$^{1}$ \and Nutan Chen$^{2}$%
  \thanks{$^{1}$M. Dawood, O. Drama Berkenkamp, and A. Gonzales Marin are with Agile Robots SE, Munich, Germany.}%
  \thanks{$^{2}$N. Chen is with LS Wiiri Innovation Center, Wuxi, China.}%
}
\begin{document}

\maketitle
\thispagestyle{empty}
\pagestyle{empty}

\begin{abstract}
Achieving human-like behavior on humanoid robots has advanced rapidly by leveraging motion-capture demonstrations rather than hand-crafted cost functions. 
Adversarial imitation supplies a style signal that lets a policy fulfill a task while adhering to a desired style, but is difficult to transfer to hardware. Whole-body tracking through reinforcement learning instead deploys reliably and imitates demonstrations faithfully, but the resulting policies are conditioned on the future reference, so they require a reference trajectory at runtime and can fail when the motion differs from the training data.
We train a humanoid policy directly with whole-body tracking rewards while restricting its actor to compact locomotion commands and robot observations. Reference motion supervises the reward and privileged critic, without supplying reference joint trajectories to the actor or requiring teacher distillation. The demonstrated style is thereby encoded in the policy through the tracking reward, so the robot reproduces the whole-body style under commanded velocities.
Trained on a single random-walk motion and commanded with velocity profiles from held-out motions, our policy reaches at least 98\% success in simulation and completes every trial on the Unitree G1. On held-out slow walking, whole-body tracking drops to 2\% success and command-conditioned distillation to 56\%, and both fail on the real robot. On the trained motions, the joint position error of our policy stays within 8\% of whole-body tracking, and it has a higher style coverage than adversarial motion priors (AMP), which fell on the real robot without a safety crane.

\end{abstract}

\section{Introduction}
Humanoid robots have shown the ability to successfully perform a variety of tasks by imitating human motions \cite{zhao2025humanoid, gu2026humanoid}. This has motivated the capture and generation of extensive motion datasets through motion-capture systems \cite{bonesstudio2025bonesseed, mahmood2019amass, harvey2020robust}. However, these datasets do not account for the physical constraints of the robot, such as its kinematic and dynamic limits \cite{xie2026kungfubot}. As a result, imitating the datasets directly through supervised learning yields policies that fail to execute reliably on the robot. Reinforcement learning (RL), on the other hand, is able to learn locomotion policies through exploration that exploits the robot's own dynamics and kinematics \cite{lee2020learning, hwangbo2019learning, margolis2024rapid, kumar2021rma, radosavovic2024real}. This has made RL a promising approach for learning stylized locomotion behaviors from motion references \cite{peng2021amp, peng2018deepmimic, yang2025omniretarget}.

Existing pipelines for learning stylized locomotion follow one of three strategies. Adversarial training provides the RL agent with a reward signal about the style \cite{peng2021amp, mu2025smp, peng2022ase, wen2025constrained}. Whole-body tracking instead combines multiple rewards to imitate the reference motion by tracking joint positions, joint velocities, and the positions and velocities of key body links \cite{xie2026kungfubot, peng2018deepmimic, liao2025beyondmimic, yang2025omniretarget}. A third strategy distills a privileged, reference-conditioned teacher into a student that observes only deployable inputs together with a history of past observations \cite{ji2024exbody2}. Each of them has a cost in reliability, in controllability, or in how the policy is trained.
\begin{figure}
    \centering
    \includegraphics[width=0.99\linewidth]{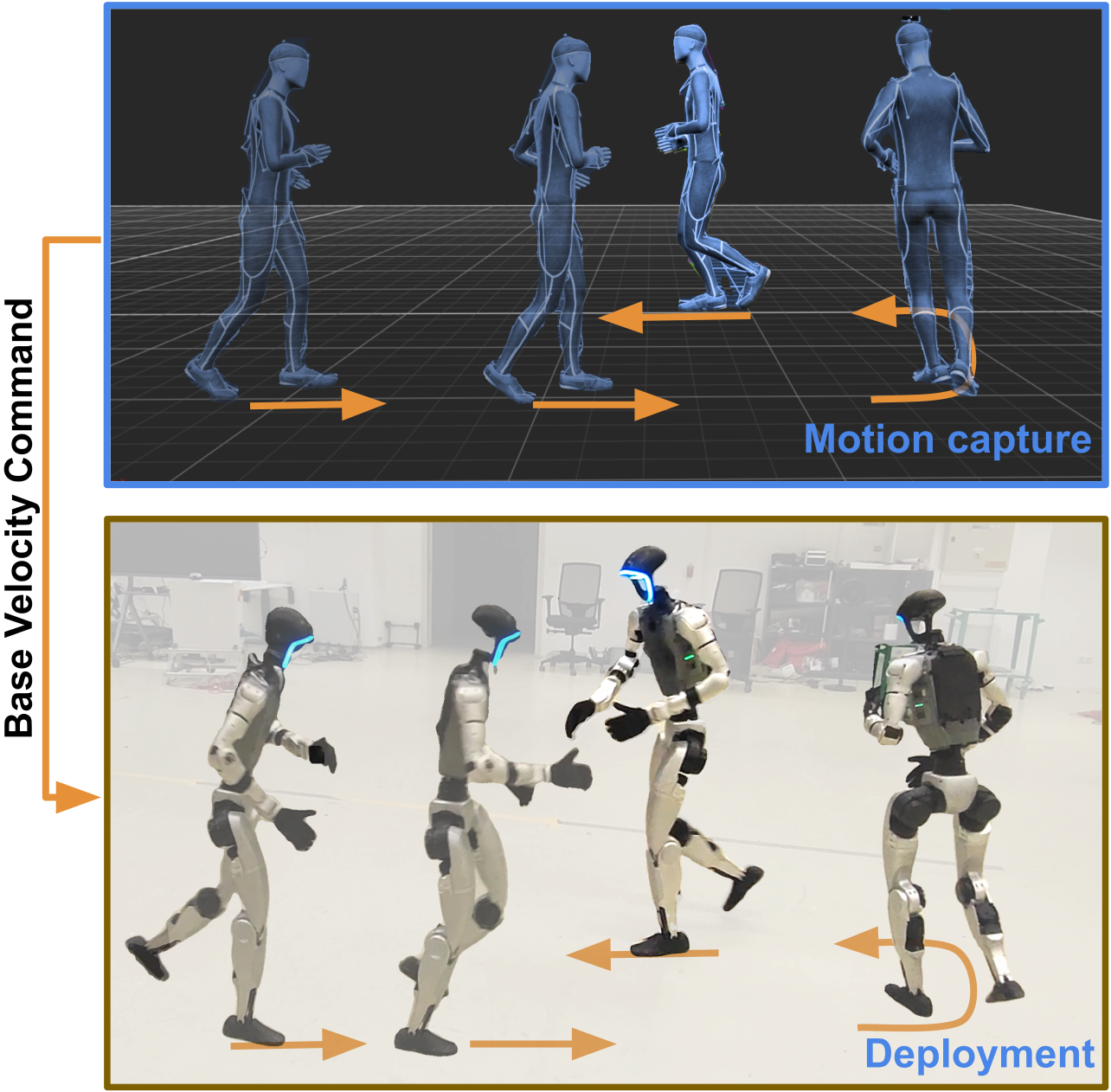}
    \caption{Our policy learns the jogging style from a motion-capture clip (top) and reproduces it on the Unitree G1 (bottom). Only the base velocity of the clip (orange) is given to the policy as a command. The reference joints and body poses are used only during training.}
    \label{fig:cover}
\end{figure}
Adversarial approaches have seen great success in computer graphics \cite{peng2021amp, peng2022ase}: beyond imitating a style, they allow the policy to be driven at runtime without a reference trajectory. However, their deployment to the real world is challenging because of the weighting between the style reward and the task reward. Emphasizing the style can lead to unsuccessful deployment on the real robot (e.g. jittery motions and stumbling), while emphasizing the task reward enables successful deployment at the cost of weaker style imitation. Whole-body tracking, on the other hand, can deploy reliably on real robots while imitating the reference motion to a great extent. However, these policies are conditioned on the future reference of joint positions and velocities, so they require the reference trajectory at runtime and the whole-body motion must come from a recorded clip. Command-conditioned distillation can give the student a low-dimensional command interface, but the student learns by imitating the actions of a reference-conditioned teacher rather than from the task reward, so its behavior is limited by what the teacher does.

To close this gap, we present a velocity-conditioned whole-body controller that deploys reliably and can be steered at runtime, without distillation from a privileged teacher. We train a tracking policy with an asymmetric actor-critic: the critic observes the full reference motion, while the actor observes only proprioception, the anchor orientation error, and a planar base velocity command. The demonstrated style is thereby encoded in the policy through the tracking reward, so the robot reproduces the whole-body style under a commanded velocity (Fig.~\ref{fig:cover}). Our key observation is that different locomotion motions often share the base velocities of the training motion while their joint trajectories differ: forward walking, sidesteps, and turning all reduce to base-level goals. A policy conditioned on the velocity command, and not on the reference joints, is therefore less dependent on the joint trajectories of the training clip and can transfer to such motions. At deployment, its inputs can be computed from the base trajectory of a motion clip or, without a clip, from a gamepad. In our experiments, we drive it with velocity profiles from held-out motion clips, so that tracking and style can be measured against ground truth. We compare our approach with all baselines on the Unitree G1, in simulation and on real hardware..

In summary, our contributions are:
\begin{itemize}
\item A velocity-conditioned whole-body tracking policy, trained directly with RL and an asymmetric actor-critic, whose actor observes only proprioception, the anchor orientation error, and a planar velocity command, without distillation from a privileged teacher.
\item Simulation and hardware results. Trained on a random walk and commanded with the velocity profiles of three held-out motions, our policy reaches at least 98\% success in simulation and completes every hardware trial, while whole-body tracking and command-conditioned distillation fail on slow walking. Compared with AMP, our policy runs on the real robot without support, while AMP fell in both attempts without a safety crane, and it tracks the reference more closely with a higher style coverage. On the trained motions, its mean joint position error stays within about $8\%$ of the reference-conditioned tracker.

\item An analysis of the policy inputs, showing that 87 to 100\% of the held-out velocity commands lie close to the training data, compared with 19 to 87\% of the reference joint frames, together with ablations on mirrored data, symmetry training, and smoothed, gamepad-like motions.
\end{itemize}
\section{Related Work}
\subsection{Humanoid Whole-Body Motion Tracking}
RL-based motion tracking has shown great success recently in imitating natural
human movements on humanoids. \cite{peng2018deepmimic} presented an RL framework
that trains policies to imitate recorded motions via a set of rewards for tracking
the positions and velocities of selected body links. \cite{yang2025omniretarget}
generated motions that preserve the interactions with the environment, including
the objects to be interacted with, so that tracking policies can learn to interact
with the environment while imitating the motions. \cite{he2025asap} added a
residual action model to account for the dynamics mismatch between simulation and
the real world and improve sim-to-real transfer. \cite{xie2026kungfubot} applied
motion filtering, feet contact masks, and adaptive rewards to enable learning of
highly dynamic motions. \cite{sleiman2026zest} combined adaptive sampling of
motions based on difficulty with a virtual assistive wrench to help stabilize the
training of athletic moves. \cite{ji2024exbody2} applied a teacher-student DAgger
\cite{ross2011reduction} with historical observations to learn the motions from a
privileged teacher, and also let a joystick command the root velocity and pose,
while the policy still receives the whole-body joint and keypoint targets. These
approaches imitate the recorded motions faithfully, but the policies require the
reference trajectory at runtime. Different from these approaches, our goal is a
policy that is controlled by a velocity command and does not need a reference
trajectory at deployment.

\subsection{Style Imitation from Demonstrations}
On the other hand, several works aimed to make the imitation controllable without
a reference trajectory at runtime. \cite{peng2021amp} utilized adversarial training
to make physically simulated characters imitate a given style without the need for
reward shaping. Rather than imitating a single clip, \cite{peng2022ase, mu2025smp} learned
reusable motion priors that encode a repertoire of skills from large unstructured
datasets, via a latent skill embedding and a score-matching (diffusion-based) prior
respectively, which a downstream controller can then use to perform new tasks.
However, \cite{wen2025constrained} showed that adversarial training requires careful
tuning of the task and style rewards to avoid sub-optimal behaviors; the authors
formulated the problem as a constrained Markov decision process and deployed it on a
real quadruped. This tuning is even harder on humanoids: \cite{cheng2024expressive}
reported that adding an adversarial style reward leads to stumbling and jittering
that is not feasible for sim-to-real transfer. \textcolor{black}{\cite{ma2025styleloco} distilled an RL velocity-tracking
teacher into a student with discriminators against both the teacher and
motion-capture data, but the style still relies on adversarial rewards that must be
balanced against the task.}

To make the imitation robust for deployment, other works used hybrid controllers
that split the body. \cite{cheng2024expressive} trained the lower body to track a
base velocity robustly while the upper body imitates a reference motion.
\cite{shi2026adversarial} applied a similar split, where the lower body tracks a
velocity command and the upper body tracks a motion, using adversarial learning
between the two policies so each stays robust to the other's movements. These
controllers deploy reliably and can be steered through the velocity command, but
the style is limited to the upper body while the lower body follows a plain
velocity-tracking objective. Unlike these approaches, we train the whole body with
the dense tracking rewards of \cite{peng2018deepmimic}, so the style is reproduced
across the whole body, and we condition the policy on the base velocity commands to
keep it steerable at deployment.

\section{Approach}
\label{sec:approach}

Our pipeline has two stages: (i) we record human motions with a motion-capture system and retarget them to the robot, and (ii) we train a whole-body tracking policy with reinforcement learning under an asymmetric actor--critic, where the actor is restricted to deployable observations conditioned on a base-velocity command. At deployment the velocity command is supplied online (e.g.\ from a game-pad) to steer the robot while it reproduces the demonstrated style.

\subsection{Motion Acquisition and Retargeting}
We record source motions with a motion-capture system and retarget them to the robot with GMR~\cite{araujo2025retargeting}, which solves an inverse-kinematics problem mapping the captured human trajectory onto the robot kinematics while respecting its joint limits. For each clip this yields a per-frame reference trajectory of joint positions and velocities and of the position, orientation, and linear/angular velocity of a set of tracked bodies, expressed in the world frame. A designated anchor body (the pelvis) defines the reference base frame.

\subsection{Problem Formulation}
We formulate whole-body tracking as a Markov decision process $(\mathcal{S}, \mathcal{A}, p, r, \gamma)$ and solve it with proximal policy optimization (PPO)~\cite{schulman2017proximal}. The policy runs at $50\,\mathrm{Hz}$
(simulation timestep $5\,\mathrm{ms}$, decimation $4$) and maximizes the discounted return $\mathbb{E}\!\left[\sum_t \gamma^t r_t\right]$.

\paragraph{Actions.}
The action $a_t \in \mathbb{R}^{n}$ is a target for the $n$ actuated joints, added to the default joint configuration and tracked by joint-level PD controllers.

\paragraph{Observations (asymmetric actor--critic).}
The critic receives privileged, simulation-only information while the actor receives only inputs available on the real robot. The critic observation comprises the full reference: anchor position and orientation error, per-body
positions and orientations, and reference base linear and angular velocity, together with proprioception. The actor observation is
\begin{equation}
x_t = \big[\, v^{\text{cmd}}_t,\; R^{\text{anchor}}_t,\; \omega_t,\;
      q_t - q_{\text{def}},\; \dot q_t - \dot q_{\text{def}},\; a_{t-1} \,\big],
\end{equation}
where $v^{\text{cmd}}_t = (v_x, v_y, \omega_z) \in \mathbb{R}^{3}$ is the reference planar base velocity and yaw rate expressed in the robot base frame (the online steering command), $R^{\text{anchor}}_t$ is the \textcolor{black}{anchor orientation error defined below}, $\omega_t$ the base angular velocity, $q_t,\dot q_t$ the joint positions/velocities from the encoders, and $a_{t-1}$ the previous action.

\begingroup\color{black}The anchor orientation error is the orientation of a target frame relative to the robot pelvis (the anchor body):
\begin{equation}
R^{\text{anchor}}_t = \big(R^{\text{robot}}_t\big)^{\top} R^{\text{target}}_t,
\label{eq:anchor_err}
\end{equation}
where $R^{\text{robot}}_t$ is the measured pelvis orientation. The actor receives the first two columns of $R^{\text{anchor}}_t$ (6 values). In training, the target is the orientation of the reference pelvis, $R^{\text{target}}_t = R^{\text{ref}}_t$.

The actor therefore never observes the reference joints or body poses. The task is given only through the base motion: the velocity command $v^{\text{cmd}}_t$ and the target orientation $R^{\text{target}}_t$. The demonstrated style is absorbed into the policy weights through the tracking reward. This is what makes the policy steerable online without a reference trajectory at runtime.\endgroup

\paragraph{Steering at deployment.}
\begingroup\color{black}At deployment, the policy needs the velocity command $v^{\text{cmd}}_t = (v_x, v_y, \omega_z)$ and the target orientation $R^{\text{target}}_t$, which can be obtained in two ways. (i) From a motion clip: $v^{\text{cmd}}_t$ is the planar velocity and yaw rate of the reference pelvis, and $R^{\text{target}}_t = R^{\text{ref}}_t$ is its orientation, as in training. Only the base trajectory is used, and the joint trajectories of the clip are never given to the policy. (ii) From a gamepad: the sticks give the velocity command, scaled to the range of velocities seen in training, and the target is an upright frame, $R^{\text{target}}_t = R_z(\psi_t)$, where $R_z(\psi)$ is a rotation by $\psi$ about the vertical axis. Its heading starts at the robot heading, $\psi_0 = \psi^{\text{robot}}_0$, and is integrated with the commanded yaw rate,
\begin{equation}
\psi_{t+1} = \psi_t + \omega_z \Delta t, \qquad \Delta t = 0.02\,\mathrm{s},
\end{equation}
so no motion clip is needed. This online target keeps the heading error, which steers the robot, and drops only the roll and pitch of the reference pelvis.\endgroup{} In our experiments (Sec.~\ref{sec:results}), we use option (i) with held-out motions, so that tracking and style can be measured against their references.

\subsection{Reward}
The reward combines exponential tracking terms of the form $w_i \exp(-\lVert e_i \rVert^2 / \sigma_i^2)$ with regularization penalties. Table~\ref{tab:rewards} lists all terms.

\begin{table}[t]
\centering
\caption{Reward terms. Tracking terms use the exponential kernel $\exp(-e^2/\sigma^2)$; penalties are quadratic. The body-tracking weights are ramped from their initial to final value during standing-start episodes
(Sec.~\ref{sec:init}); elsewhere they take the final value.}
\label{tab:rewards}
\begin{tabular}{llrr}
\toprule
Term & Error $e$ & Weight $w$ & $\sigma$ \\
\midrule
\multicolumn{4}{l}{\emph{Tracking (exponential)}}\\
Anchor position        & global anchor pos.\ error      & $1.0$          & $0.3$  \\
Anchor orientation     & global anchor ori.\ error      & $1.0$          & $0.4$  \\
Body position          & per-body rel.\ pos.\ error     & $0.7\!\to\!1.0$ & $0.3$  \\
Body orientation       & per-body rel.\ ori.\ error     & $0.7\!\to\!1.0$ & $0.4$  \\
Body linear velocity   & global body lin.\ vel.\ error  & $0.7\!\to\!1.0$ & $1.0$  \\
Body angular velocity  & global body ang.\ vel.\ error  & $0.7\!\to\!1.0$ & $3.14$ \\
Lateral deviation      & lateral base drift             & $0.5$          & $0.2$  \\
\midrule
\multicolumn{4}{l}{\emph{Regularization (penalty)}}\\
Action rate            & $\lVert a_t - a_{t-1}\rVert^2$  & $-0.1$         & --- \\
Joint acceleration     & $\lVert \ddot q_t \rVert^2$     & $-10^{-5}$     & --- \\
Joint-limit violation  & soft-limit excess              & $-2.0$         & --- \\
Self-collision         & pelvis self-contact            & $-5.0$         & --- \\
\bottomrule
\end{tabular}
\end{table}

\subsection{Initialization and Curriculum}
\label{sec:init}

\paragraph{Initialization and sampling.}
Episodes are initialized by reference-state initialization (RSI)~\cite{peng2018deepmimic}, resetting the robot to a reference frame of the clip (with small pose and velocity perturbations) so the policy learns all phases of the motion in parallel rather than only from the start. We select the start frame with the failure-based adaptive sampler of~\cite{liao2025beyondmimic}, which partitions the clip into one-second bins and concentrates episodes on the bins where the policy terminates most often, with a uniform floor that preserves exploration over all bins. This focuses training on the segments the policy struggles with while still revisiting the full motion.

\paragraph{Standing-start stabilization.}
The real robot always begins from a neutral standing pose, a state RSI never produces. To close this gap we initialize a fraction $p_{\text{stand}} = 0.3$ of episodes from the robot's default \emph{standing} state
while anchoring the reference at the first motion frame. These standing-start episodes are eased into tracking in two phases governed by two parameters:
\begin{itemize}
  \item \textbf{Stabilization} ($N_{\text{stab}}=30$ steps): the reference is held frozen at the first frame so the robot can settle from standing into the initial pose before the motion begins to play.
  \item \textbf{Transition} ($N_{\text{trans}}=60$ steps): the reference then begins to advance while the body-tracking reward weights ramp linearly from $0.7$ to $1.0$, smoothly handing control from ``stabilize'' to ``fully track.''
\end{itemize}
Formally, a per-episode progress signal $\phi_t = \mathrm{clip}\!\big((t - N_{\text{stab}})/N_{\text{trans}},\,0,\,1\big)$ scales the body-tracking weights as $w = w_{\text{init}} + (w_{\text{final}}-w_{\text{init}})\,\phi_t$. For ordinary RSI episodes $\phi_t \equiv 1$, so these two parameters affect only the standing-start subset. We use this procedure so that the policy is trained on starts from rest, as on the real robot.


\section{Results}
\label{sec:results}

We compare our policy with three baselines in simulation and on the Unitree G1. We first evaluate all methods on the motions they were trained on and on held-out motions, in simulation (Sec.~\ref{sec:res_sim}) and on the real robot (Sec.~\ref{sec:res_real}). We then compare the inputs of our policy and of the whole-body tracking policy with the training data (Sec.~\ref{sec:res_coverage}). Finally, we present two ablations, each in simulation and on the real robot: mirrored training data and symmetry training (Sec.~\ref{sec:res_sym}), and training on smoothed motions (Sec.~\ref{sec:res_smooth}).

\subsection{Experimental Setup}
\label{sec:res_setup}

\begin{table}[t]
\centering
\scriptsize
\setlength{\tabcolsep}{3pt}
\caption{Inputs and training of the compared methods. All methods also observe proprioception. Command: planar velocity command. Real robot: free = without support. Ours is the only method trained with the tracking reward that needs neither the reference joints nor a teacher.}
\label{tab:baselines}
\begin{tabular}{lccccc}
\toprule
Method & Ref.\ joints & Command & History & Supervision & Real robot \\
\midrule
Teacher~\cite{liao2025beyondmimic} & next step & no & no & tracking reward & free \\
Distill & no & yes & 10 steps & Teacher actions & free \\
AMP~\cite{peng2021amp} & no & yes & no & AMP reward & crane \\
Ours & no & yes & no & tracking reward & free \\
\bottomrule
\end{tabular}
\end{table}

\paragraph{Baselines.}
Table~\ref{tab:baselines} summarizes the compared methods.
\begin{itemize}
  \item \emph{Teacher}: the whole-body tracking policy of BeyondMimic~\cite{liao2025beyondmimic}, using its implementation in mjlab. At each step $t$ it observes the reference joint positions and velocities of the next step $t+1$.
  \item \emph{Distill}: a command-conditioned distillation baseline that follows the teacher-student training of ExBody2~\cite{ji2024exbody2}. The student has the same inputs as our policy together with a history of the last 10 observations, and it does not observe the reference joints. It is trained from the final Teacher policy with DAgger~\cite{ross2011reduction}: the student is rolled out in simulation, the Teacher computes its action $\hat a_t$ for every visited state, and the student is updated to minimize the mean squared error $\lVert a_t - \hat a_t \rVert^2$ between its own action $a_t$ and the Teacher action. Training continues until the loss stops changing, which took about 20k iterations. Unlike ExBody2, where the student also receives the motion targets, our student receives only the velocity command, so Distill is a teacher-student version of our policy.
  \item \emph{AMP}~\cite{peng2021amp}: adversarial motion priors. A discriminator is trained to tell apart state transitions of the reference motion from those of the policy, and its output is used as a style reward. The policy maximizes this style reward together with a task reward for following the velocity command, so it does not track the reference frame by frame.
\end{itemize}
All methods are trained with the same number of parallel environments, the same domain randomization, and the same number of iterations. The only exception is Distill, which is trained from the final Teacher policy as described above.

\paragraph{Motions and protocol.}
In distribution, each policy is trained and evaluated on the same motion: random walk, backward, side, slow, and jog in simulation, and random walk, slow, and jog on the real robot. For generalization, the policies are trained on the random walk and evaluated on the held-out backward, side, and slow walking motions. The velocity commands are taken from the held-out clips, so the reference joints are available to measure tracking and style. In simulation, each motion is rolled out in 128 parallel environments for two passes of the clip. On the real robot, each motion is run in three trials. Trials that were interrupted by problems unrelated to the policy, such as communication disconnections, were repeated. Without support, AMP fell in both of two attempts on the real robot, so its trials were run on a safety crane.

\paragraph{Metrics.}
\begin{itemize}
  \item \emph{Success}: in simulation, the percentage of environments that do not terminate during the two passes of the clip. The tracking-based methods terminate when the anchor height, the anchor orientation, or the ankle and wrist heights deviate too far from the reference ($0.25\,\mathrm{m}$ for the heights). AMP terminates on falls (tilt above $70^\circ$ or pelvis height below $0.5\,\mathrm{m}$); we checked offline that the AMP rollouts also pass the tracking terminations on every motion. On the real robot, success is the number of successful trials out of three.
  \item \emph{MJPE} and \emph{MJVE}: the mean absolute joint position and velocity errors per joint, averaged over the 29 joints and all frames.
  \item \emph{MMD} (minimum matching distance) and \emph{COV} (coverage): compare the distribution of the generated joint velocities with the reference. Both sets are projected onto the first three principal components of the reference joint velocities and then standardized with the reference statistics. MMD is the mean nearest-neighbor distance between the two sets in both directions, and COV is the share of reference samples that are the nearest neighbor of at least one generated sample.
\end{itemize}
All methods are evaluated against the same motion recordings; command-conditioned methods receive the corresponding planar velocity profiles and start from the same pose.

\subsection{Comparison in Simulation}
\label{sec:res_sim}

\begin{table}[t]
\centering
\scriptsize
\setlength{\tabcolsep}{2.5pt}
\caption{Simulation, in distribution: each policy is trained and evaluated on the same motion. Success over 128 environments, MJPE in deg, MJVE in rad/s. Best values in bold. }
\label{tab:sim_id}
\begin{tabular}{llccccc}
\toprule
Motion & Method & Succ.$\uparrow$ & MJPE$\downarrow$ & MJVE$\downarrow$ & MMD$\downarrow$ & COV$\uparrow$ \\
\midrule
\multirow{4}{*}{Random walk} & AMP & \textbf{100\%} & 6.64 & 0.63 & 0.14 & 0.39 \\
 & Teacher & \textbf{100\%} & \textbf{2.77} & \textbf{0.32} & 0.11 & 0.46 \\
 & Distill & \textbf{100\%} & 2.95 & 0.34 & 0.11 & 0.46 \\
 & Ours & \textbf{100\%} & 3.07 & 0.34 & \textbf{0.10} & \textbf{0.47} \\
\midrule
\multirow{4}{*}{Backward} & AMP & \textbf{100\%} & 5.06 & 0.46 & 0.13 & 0.29 \\
 & Teacher & \textbf{100\%} & \textbf{2.64} & \textbf{0.23} & \textbf{0.10} & 0.38 \\
 & Distill & \textbf{100\%} & 2.86 & 0.24 & \textbf{0.10} & \textbf{0.40} \\
 & Ours & \textbf{100\%} & 2.90 & 0.26 & \textbf{0.10} & 0.38 \\
\midrule
\multirow{4}{*}{Side} & AMP & \textbf{100\%} & 2.85 & 0.25 & 0.17 & 0.33 \\
 & Teacher & \textbf{100\%} & \textbf{2.41} & \textbf{0.16} & \textbf{0.14} & 0.42 \\
 & Distill & \textbf{100\%} & 2.53 & 0.18 & 0.15 & \textbf{0.44} \\
 & Ours & \textbf{100\%} & 2.52 & 0.19 & 0.15 & 0.42 \\
\midrule
\multirow{4}{*}{Slow} & AMP & \textbf{100\%} & 4.85 & 0.15 & 0.11 & 0.24 \\
 & Teacher & \textbf{100\%} & \textbf{2.66} & \textbf{0.13} & \textbf{0.09} & \textbf{0.29} \\
 & Distill & \textbf{100\%} & 2.68 & 0.14 & 0.10 & \textbf{0.29} \\
 & Ours & \textbf{100\%} & 2.78 & 0.15 & 0.12 & 0.28 \\
\midrule
\multirow{4}{*}{Jog} & AMP & \textbf{100\%} & 6.58 & 1.10 & 0.24 & 0.17 \\
 & Teacher & \textbf{100\%} & \textbf{3.30} & \textbf{0.53} & 0.13 & 0.34 \\
 & Distill & \textbf{100\%} & 3.39 & 0.54 & 0.13 & \textbf{0.35} \\
 & Ours & \textbf{100\%} & 3.62 & 0.56 & \textbf{0.12} & 0.33 \\
\midrule
\multirow{4}{*}{Avg.} & AMP & \textbf{100\%} & 5.20 & 0.52 & 0.16 & 0.28 \\
 & Teacher & \textbf{100\%} & \textbf{2.76} & \textbf{0.28} & \textbf{0.11} & 0.38 \\
 & Distill & \textbf{100\%} & 2.88 & 0.29 & \textbf{0.11} & \textbf{0.39} \\
 & Ours & \textbf{100\%} & 2.98 & 0.30 & 0.12 & 0.38 \\
\bottomrule
\end{tabular}
\end{table}

\begin{table}[t]
\centering
\scriptsize
\setlength{\tabcolsep}{2.5pt}
\caption{Simulation, generalization: policies trained on the random walk and evaluated on held-out motions. Averages are over the motions with at least one success. Bold: best value among the methods \textcolor{black}{with at least 90\% success}.}
\label{tab:sim_gen}
\begin{tabular}{llccccc}
\toprule
Motion & Method & Succ.$\uparrow$ & MJPE$\downarrow$ & MJVE$\downarrow$ & MMD$\downarrow$ & COV$\uparrow$ \\
\midrule
\multirow{4}{*}{Backward} & AMP & \textbf{100\%} & 5.48 & 0.45 & \textbf{0.12} & 0.29 \\
 & Teacher & 98\% & \textbf{3.27} & \textbf{0.35} & 0.14 & \textbf{0.36} \\
 & Distill & 99\% & 4.21 & 0.43 & 0.18 & 0.32 \\
 & Ours & \textbf{100\%} & 4.05 & 0.41 & 0.14 & 0.34 \\
\midrule
\multirow{4}{*}{Side} & AMP & \textbf{100\%} & 4.88 & 0.30 & 0.18 & 0.33 \\
 & Teacher & \textbf{100\%} & \textbf{3.01} & \textbf{0.24} & \textbf{0.17} & \textbf{0.41} \\
 & Distill & 98\% & 3.58 & 0.29 & \textbf{0.17} & 0.40 \\
 & Ours & \textbf{100\%} & 3.34 & 0.27 & \textbf{0.17} & 0.39 \\
\midrule
\multirow{4}{*}{Slow} & AMP & \textbf{100\%} & \textbf{5.65} & \textbf{0.15} & \textbf{0.12} & 0.19 \\
 & Teacher & 2\% & 4.71 & 0.40 & 0.30 & 0.25 \\
 & Distill & 56\% & 6.88 & 0.48 & 0.42 & 0.23 \\
 & Ours & 98\% & 6.20 & 0.43 & 0.35 & \textbf{0.25} \\
\midrule
\multirow{4}{*}{Avg.} & AMP & \textbf{100\%} & 5.34 & \textbf{0.30} & \textbf{0.14} & 0.27 \\
 & Teacher & 67\% & 3.66 & 0.33 & 0.20 & 0.34 \\
 & Distill & 85\% & 4.89 & 0.40 & 0.26 & 0.32 \\
 & Ours & 99\% & \textbf{4.53} & 0.37 & 0.22 & \textbf{0.32} \\
\bottomrule
\end{tabular}
\end{table}

\paragraph{In distribution.}
All methods succeed on every motion (Table~\ref{tab:sim_id}), so the comparison is about tracking and style. The Teacher tracks best, with an average MJPE of $2.76^\circ$ and MJVE of $0.28\,\mathrm{rad/s}$. Distill and our policy follow with $2.88^\circ$ and $2.98^\circ$, which is $4.3\%$ and $8.0\%$ above the Teacher, although our policy has no observation history and no reference joint input. AMP, which is trained for style rather than for tracking, has about twice the error ($5.20^\circ$). In style, our policy is close to the Teacher (MMD $0.12$ vs.\ $0.11$, COV $0.38$ for both). Compared with AMP, it lowers MMD by $26\%$ and raises COV by $32\%$, so under the same commands it reproduces the reference more closely in both tracking and style.\footnotetext[1]{AMP ran on a safety crane. Without support, it fell in both of two attempts, so we used the crane to measure its tracking and style without damaging the robot. The crane only stops a fall and does not hold the robot while walking.}\setcounter{footnote}{1}

\paragraph{Generalization.}
On backward and side walking all methods succeed in at least $98\%$ of the environments (Table~\ref{tab:sim_gen}). The held-out slow walking separates them: success is $2\%$ for the Teacher, $56\%$ for Distill, $98\%$ for our policy, and $100\%$ for AMP. Our policy is therefore the only tracking-based method with at least $98\%$ success on every held-out motion. On backward and side walking, the Teacher tracks best (MJPE $3.27^\circ$ and $3.01^\circ$), followed by our policy ($4.05^\circ$ and $3.34^\circ$), Distill ($4.21^\circ$ and $3.58^\circ$), and AMP ($5.48^\circ$ and $4.88^\circ$), so our policy also tracks the held-out references more closely than AMP. Transfer comes with a lower fidelity: on slow walking, the MJPE of our random-walk policy is $6.20^\circ$, compared with $2.78^\circ$ for our policy trained on slow walking. Our policy has a higher COV than AMP on all three held-out motions, while AMP has a lower MMD on slow walking ($0.12$ vs.\ $0.35$). On slow walking, the metrics of the Teacher and Distill come only from the $2$ and $72$ of $128$ environments that succeeded.

\paragraph{Tracking versus generalization.}
The Teacher tracks best wherever it succeeds, but it fails on the held-out slow walking. Distill, which is distilled from the Teacher, fails less often but remains well below our policy. Our policy gives up some tracking accuracy and in return reaches at least $98\%$ success on every held-out motion. Since our goal is a policy that keeps working on motions it was not trained on, we treat success on held-out motions as the main measure and tracking error as secondary.

\subsection{Comparison on the Real Robot}
\label{sec:res_real}

\begin{table}[t]
\centering
\scriptsize
\setlength{\tabcolsep}{2.5pt}
\caption{Real robot, in distribution (jog and slow: per-motion policies, random walk: random-walk policy). Success out of 3 trials; values are mean $\pm$ std over the successful trials.}
\label{tab:real_id}
\begin{tabular}{llccccc}
\toprule
Motion & Method & Succ.$\uparrow$ & MJPE$\downarrow$ & MJVE$\downarrow$ & MMD$\downarrow$ & COV$\uparrow$ \\
\midrule
\multirow{4}{*}{Jog} & \textcolor{black}{AMP\footnotemark[1]} & 3/3 & 7.53{\tiny$\pm$0.12} & 1.23{\tiny$\pm$0.03} & 0.26{\tiny$\pm$0.01} & 0.20{\tiny$\pm$0.01} \\
 & Teacher & \textbf{3/3} & \textbf{3.78}{\tiny$\pm$0.03} & \textbf{0.63}{\tiny$\pm$0.00} & \textbf{0.16}{\tiny$\pm$0.00} & \textbf{0.30}{\tiny$\pm$0.00} \\
 & Distill & \textbf{3/3} & 4.53{\tiny$\pm$0.13} & 0.68{\tiny$\pm$0.01} & 0.17{\tiny$\pm$0.00} & 0.29{\tiny$\pm$0.01} \\
 & Ours & \textbf{3/3} & 4.00{\tiny$\pm$0.01} & 0.66{\tiny$\pm$0.00} & \textbf{0.16}{\tiny$\pm$0.00} & 0.27{\tiny$\pm$0.00} \\
\midrule
\multirow{4}{*}{Slow} & \textcolor{black}{AMP\footnotemark[1]} & 3/3 & 5.59{\tiny$\pm$0.22} & 0.21{\tiny$\pm$0.01} & 0.21{\tiny$\pm$0.02} & 0.31{\tiny$\pm$0.02} \\
 & Teacher & \textbf{3/3} & \textbf{2.77}{\tiny$\pm$0.01} & \textbf{0.13}{\tiny$\pm$0.00} & \textbf{0.10}{\tiny$\pm$0.00} & \textbf{0.39}{\tiny$\pm$0.00} \\
 & Distill & \textbf{3/3} & 2.90{\tiny$\pm$0.04} & 0.15{\tiny$\pm$0.00} & 0.12{\tiny$\pm$0.00} & 0.37{\tiny$\pm$0.01} \\
 & Ours & \textbf{3/3} & 2.96{\tiny$\pm$0.01} & 0.14{\tiny$\pm$0.00} & 0.12{\tiny$\pm$0.00} & 0.37{\tiny$\pm$0.00} \\
\midrule
\multirow{4}{*}{Random walk} & \textcolor{black}{AMP\footnotemark[1]} & 3/3 & 6.65{\tiny$\pm$0.02} & 0.66{\tiny$\pm$0.01} & 0.15{\tiny$\pm$0.00} & 0.39{\tiny$\pm$0.00} \\
 & Teacher & \textbf{3/3} & \textbf{3.03}{\tiny$\pm$0.01} & \textbf{0.35}{\tiny$\pm$0.00} & 0.14{\tiny$\pm$0.00} & \textbf{0.43}{\tiny$\pm$0.00} \\
 & Distill & \textbf{3/3} & 3.33{\tiny$\pm$0.01} & 0.37{\tiny$\pm$0.00} & \textbf{0.13}{\tiny$\pm$0.00} & \textbf{0.43}{\tiny$\pm$0.00} \\
 & Ours & \textbf{3/3} & 3.68{\tiny$\pm$0.04} & 0.41{\tiny$\pm$0.00} & \textbf{0.13}{\tiny$\pm$0.00} & 0.42{\tiny$\pm$0.00} \\
\midrule
\multirow{4}{*}{Avg.} & \textcolor{black}{AMP\footnotemark[1]} & 100\% & 6.59 & 0.70 & 0.21 & 0.30 \\
 & Teacher & \textbf{100\%} & \textbf{3.19} & \textbf{0.37} & \textbf{0.13} & \textbf{0.37} \\
 & Distill & \textbf{100\%} & 3.59 & 0.40 & 0.14 & \textbf{0.37} \\
 & Ours & \textbf{100\%} & 3.55 & 0.40 & 0.14 & 0.36 \\
\bottomrule
\end{tabular}
\end{table}

\begin{table}[t]
\centering
\scriptsize
\setlength{\tabcolsep}{2.5pt}
\caption{Real robot, generalization: policies trained on the random walk. Metrics are computed on the successful trials only; --: none succeeded. Averages are over the motions with at least one successful trial. Bold: best value among the methods that complete all trials.}
\label{tab:real_gen}
\begin{tabular}{llccccc}
\toprule
Motion & Method & Succ.$\uparrow$ & MJPE$\downarrow$ & MJVE$\downarrow$ & MMD$\downarrow$ & COV$\uparrow$ \\
\midrule
\multirow{4}{*}{Backward} & \textcolor{black}{AMP\footnotemark[1]} & 3/3 & 5.46{\tiny$\pm$0.02} & 0.42{\tiny$\pm$0.00} & 0.13{\tiny$\pm$0.00} & 0.29{\tiny$\pm$0.00} \\
 & Teacher & \textbf{3/3} & \textbf{3.51}{\tiny$\pm$0.03} & \textbf{0.35}{\tiny$\pm$0.01} & \textbf{0.12}{\tiny$\pm$0.00} & \textbf{0.42}{\tiny$\pm$0.00} \\
 & Distill & \textbf{3/3} & 4.10{\tiny$\pm$0.02} & 0.41{\tiny$\pm$0.01} & 0.13{\tiny$\pm$0.00} & 0.41{\tiny$\pm$0.01} \\
 & Ours & \textbf{3/3} & 4.09{\tiny$\pm$0.02} & 0.40{\tiny$\pm$0.01} & \textbf{0.12}{\tiny$\pm$0.00} & 0.40{\tiny$\pm$0.01} \\
\midrule
\multirow{4}{*}{Side} & \textcolor{black}{AMP\footnotemark[1]} & 3/3 & 4.82{\tiny$\pm$0.03} & 0.27{\tiny$\pm$0.01} & 0.20{\tiny$\pm$0.01} & 0.34{\tiny$\pm$0.01} \\
 & Teacher & \textbf{3/3} & \textbf{2.88}{\tiny$\pm$0.02} & \textbf{0.22}{\tiny$\pm$0.00} & \textbf{0.15}{\tiny$\pm$0.00} & \textbf{0.46}{\tiny$\pm$0.00} \\
 & Distill & \textbf{3/3} & 3.37{\tiny$\pm$0.01} & 0.23{\tiny$\pm$0.00} & \textbf{0.15}{\tiny$\pm$0.00} & 0.41{\tiny$\pm$0.01} \\
 & Ours & \textbf{3/3} & 3.29{\tiny$\pm$0.02} & 0.23{\tiny$\pm$0.00} & \textbf{0.15}{\tiny$\pm$0.00} & 0.43{\tiny$\pm$0.01} \\
\midrule
\multirow{4}{*}{Slow} & \textcolor{black}{AMP\footnotemark[1]} & 3/3 & \textbf{5.58}{\tiny$\pm$0.12} & \textbf{0.19}{\tiny$\pm$0.01} & \textbf{0.18}{\tiny$\pm$0.00} & 0.30{\tiny$\pm$0.01} \\
 & Teacher & 0/3 & -- & -- & -- & -- \\
 & Distill & 1/3 & 6.14 & 0.37 & 0.36 & 0.32 \\
 & Ours & \textbf{3/3} & 5.96{\tiny$\pm$0.14} & 0.39{\tiny$\pm$0.02} & 0.37{\tiny$\pm$0.03} & \textbf{0.34}{\tiny$\pm$0.01} \\
\midrule
\multirow{4}{*}{Avg.} & \textcolor{black}{AMP\footnotemark[1]} & 100\% & 5.29 & \textbf{0.29} & \textbf{0.17} & 0.31 \\
 & Teacher & 67\% & 3.20 & 0.28 & 0.14 & 0.44 \\
 & Distill & 78\% & 4.54 & 0.34 & 0.22 & 0.38 \\
 & Ours & \textbf{100\%} & \textbf{4.45} & 0.34 & 0.22 & \textbf{0.39} \\
\bottomrule
\end{tabular}
\end{table}

Without support, AMP fell in both of two attempts, so its trials were run on a safety crane. The crane allowed us to measure the tracking and style of AMP without damaging the robot. AMP completed all three trials on the crane on every motion, but since the crane would stop a fall, its success (marked with footnote 1 in the tables) is not directly comparable with the other three methods, which ran without support.

\paragraph{In distribution.}
The Teacher, Distill, and our policy complete all three trials on every motion (Table~\ref{tab:real_id}), and the ranking from simulation holds. The Teacher tracks best (MJPE $3.19^\circ$), our policy and Distill have similar means ($3.55^\circ$ and $3.59^\circ$), and AMP has the largest error ($6.59^\circ$), as in simulation. On the same motions, the MJPE of our policy on hardware is $0.2^\circ$ to $0.6^\circ$ higher than in simulation. In style, our policy (MMD $0.14$, COV $0.36$) is close to the Teacher ($0.13$, $0.37$) and Distill ($0.14$, $0.37$).

\paragraph{Generalization.}
Our policy completes all three trials on backward, side, and slow walking (Table~\ref{tab:real_gen}), and it is the only method to do so \textcolor{black}{without support}. The Teacher fails all three trials on slow walking and Distill completes one, which matches the simulation results. On backward and side walking, the Teacher tracks best (MJPE $3.51^\circ$ and $2.88^\circ$), our policy ($4.09^\circ$ and $3.29^\circ$) is similar to Distill ($4.10^\circ$ and $3.37^\circ$), and AMP has the largest error ($5.46^\circ$ and $4.82^\circ$). Our policy has a higher COV than AMP on all held-out motions and a lower MMD on backward and side walking. On slow walking, AMP has a lower MMD ($0.18$ vs.\ $0.37$), but it ran on the crane.

\subsection{Input Coverage}
\label{sec:res_coverage}

\begin{figure}[t]
\centering
\includegraphics[width=\columnwidth]{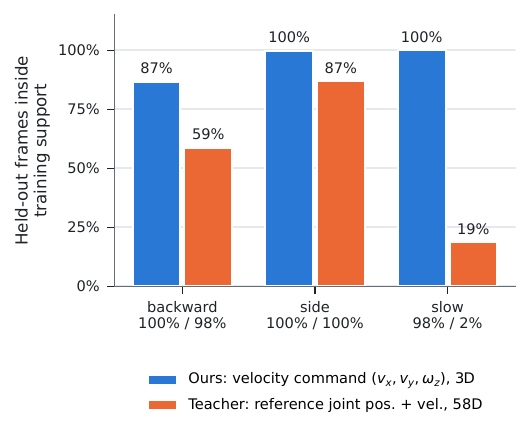}
\caption{Share of held-out frames that lie inside the support of the random-walk training data, for the input of our policy (velocity command) and of the Teacher (reference joint positions and velocities). The numbers under each motion are the simulation success of ours / Teacher. The velocity command of our policy keeps $87$ to $100\%$ of the held-out frames inside the training support, which matches its success on all of them.}
\label{fig:coverage}
\end{figure}

\begin{figure}[t]
\centering
\includegraphics[width=\columnwidth]{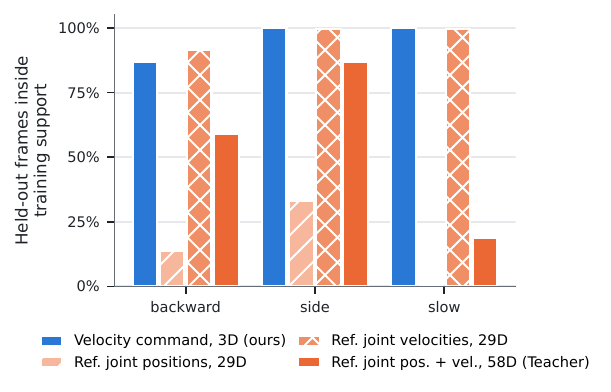}
\caption{Share of held-out frames inside the support of the training data, with the Teacher input split into reference joint positions and reference joint velocities. The reference joint positions, which our policy does not observe, are where the Teacher input leaves the training support on slow walking.}
\label{fig:coverage_space}
\end{figure}

\begin{figure*}[t]
\centering
\includegraphics[width=\textwidth]{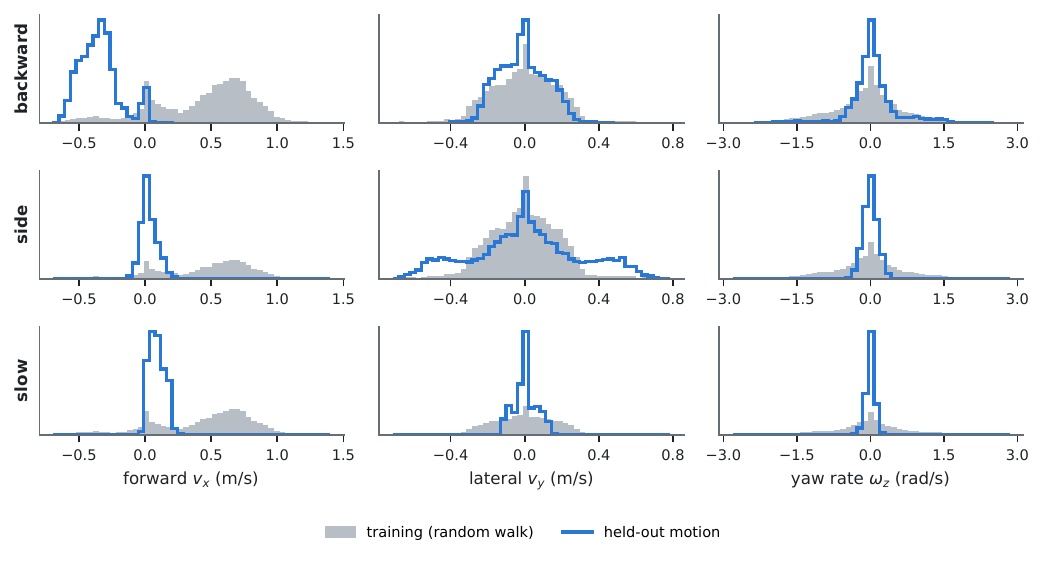}
\caption{Distribution of the velocity command in the training data (random walk, gray) and in each held-out motion (blue). The small forward velocities of slow and side walking, on which our policy succeeds, are covered by the random walk. \textcolor{black}{Backward walking falls into the smaller negative-velocity part of the random walk.}}
\label{fig:cmd_hist}
\end{figure*}

To understand why our policy keeps working on the held-out motions while the Teacher fails on slow walking, we compare the inputs of both policies with the training data. Our policy receives the planar velocity command $(v_x, v_y, \omega_z)$, while the Teacher receives the reference joint positions and velocities (58 dimensions). We count a held-out frame as inside the support of the training motion if its nearest random-walk frame is not farther than the distance between random-walk frames that are more than $2\,\mathrm{s}$ apart (95th percentile of these distances). This threshold is computed separately for each input space. The analysis uses only the motion clips and no policy.

Fig.~\ref{fig:coverage} shows the result. For the velocity command, $87\%$, $100\%$, and $100\%$ of the backward, side, and slow frames are inside the support. For the Teacher input, these values are $59\%$, $87\%$, and $19\%$. The largest gap is on slow walking, which is also where the Teacher fails. Fig.~\ref{fig:coverage_space} shows that this gap comes from the joint positions: none of the slow-walking frames are inside the support for the reference joint positions, while $99\%$ are inside for the reference joint velocities. Fig.~\ref{fig:cmd_hist} shows the velocity commands. Slow and side walking use small forward velocities that the random walk covers. \textcolor{black}{Backward walking uses negative forward velocities, which make up only $14\%$ of the random walk but span the same range, so $87\%$ of the backward frames still have a close training frame.}

\subsection{Ablation: Mirrored Data and Symmetry}
\label{sec:res_sym}

\begin{table}[t]
\centering
\scriptsize
\setlength{\tabcolsep}{2.5pt}
\caption{Simulation ablation on the random walk: B0 none, B1 mirrored training motion, B2 symmetry loss and data augmentation, B3 B1 and B2 combined, evaluated on the plain and the mirrored motion. Averages are over the motions with at least one success. Bold: best value among the methods \textcolor{black}{with at least 90\% success}.}
\label{tab:sim_sym}
\begin{tabular}{llccccc}
\toprule
Motion & Method & Succ.$\uparrow$ & MJPE$\downarrow$ & MJVE$\downarrow$ & MMD$\downarrow$ & COV$\uparrow$ \\
\midrule
\multirow{4}{*}{Plain} & B0 & \textbf{100\%} & 3.07 & \textbf{0.34} & \textbf{0.10} & \textbf{0.47} \\
 & B1 & \textbf{100\%} & 3.09 & 0.35 & \textbf{0.10} & \textbf{0.47} \\
 & B2 & \textbf{100\%} & \textbf{3.06} & \textbf{0.34} & \textbf{0.10} & \textbf{0.47} \\
 & B3 & \textbf{100\%} & 3.17 & 0.35 & \textbf{0.10} & \textbf{0.47} \\
\midrule
\multirow{4}{*}{Mirrored} & B0 & 54\% & 6.14 & 0.80 & 0.20 & 0.40 \\
 & B1 & \textbf{100\%} & 3.14 & 0.35 & \textbf{0.10} & 0.47 \\
 & B2 & \textbf{100\%} & \textbf{3.06} & \textbf{0.34} & \textbf{0.10} & \textbf{0.48} \\
 & B3 & \textbf{100\%} & 3.17 & 0.35 & \textbf{0.10} & 0.47 \\
\midrule
\multirow{4}{*}{Avg.} & B0 & 77\% & 4.61 & 0.57 & 0.15 & 0.43 \\
 & B1 & \textbf{100\%} & 3.12 & 0.35 & \textbf{0.10} & \textbf{0.47} \\
 & B2 & \textbf{100\%} & \textbf{3.06} & \textbf{0.34} & \textbf{0.10} & \textbf{0.47} \\
 & B3 & \textbf{100\%} & 3.17 & 0.35 & \textbf{0.10} & \textbf{0.47} \\
\bottomrule
\end{tabular}
\end{table}

\begin{table}[t]
\centering
\scriptsize
\setlength{\tabcolsep}{2.5pt}
\caption{Real-robot ablation on the random walk: B0 none, B1 mirrored training motion, B2 symmetry loss and data augmentation, B3 B1 and B2 combined. Metrics are computed on the successful trials only; --: none succeeded. Averages are over the motions with at least one successful trial. Bold: best value among the methods that complete all trials.}
\label{tab:real_sym}
\begin{tabular}{llccccc}
\toprule
Motion & Method & Succ.$\uparrow$ & MJPE$\downarrow$ & MJVE$\downarrow$ & MMD$\downarrow$ & COV$\uparrow$ \\
\midrule
\multirow{4}{*}{Plain} & B0 & \textbf{3/3} & 3.68{\tiny$\pm$0.04} & 0.41{\tiny$\pm$0.00} & 0.13{\tiny$\pm$0.00} & 0.42{\tiny$\pm$0.00} \\
 & B1 & \textbf{3/3} & 3.54{\tiny$\pm$0.03} & 0.41{\tiny$\pm$0.01} & \textbf{0.11}{\tiny$\pm$0.00} & \textbf{0.49}{\tiny$\pm$0.01} \\
 & B2 & \textbf{3/3} & \textbf{3.44}{\tiny$\pm$0.01} & \textbf{0.39}{\tiny$\pm$0.00} & \textbf{0.11}{\tiny$\pm$0.00} & 0.47{\tiny$\pm$0.00} \\
 & B3 & \textbf{3/3} & 3.56{\tiny$\pm$0.01} & \textbf{0.39}{\tiny$\pm$0.00} & \textbf{0.11}{\tiny$\pm$0.00} & 0.45{\tiny$\pm$0.01} \\
\midrule
\multirow{4}{*}{Mirrored} & B0 & 0/3 & -- & -- & -- & -- \\
 & B1 & \textbf{3/3} & 3.56{\tiny$\pm$0.01} & 0.41{\tiny$\pm$0.00} & \textbf{0.11}{\tiny$\pm$0.00} & \textbf{0.48}{\tiny$\pm$0.01} \\
 & B2 & \textbf{3/3} & \textbf{3.42}{\tiny$\pm$0.01} & \textbf{0.39}{\tiny$\pm$0.00} & \textbf{0.11}{\tiny$\pm$0.00} & 0.47{\tiny$\pm$0.00} \\
 & B3 & \textbf{3/3} & 3.57{\tiny$\pm$0.01} & 0.40{\tiny$\pm$0.00} & \textbf{0.11}{\tiny$\pm$0.00} & 0.45{\tiny$\pm$0.01} \\
\midrule
\multirow{4}{*}{Avg.} & B0 & 50\% & 3.68 & 0.41 & 0.13 & 0.42 \\
 & B1 & \textbf{100\%} & 3.55 & 0.41 & \textbf{0.11} & \textbf{0.48} \\
 & B2 & \textbf{100\%} & \textbf{3.43} & \textbf{0.39} & \textbf{0.11} & 0.47 \\
 & B3 & \textbf{100\%} & 3.56 & \textbf{0.39} & \textbf{0.11} & 0.45 \\
\bottomrule
\end{tabular}
\end{table}

We train four variants of the random-walk policy. B0 is trained on the original random walk only, and is the policy used in the main comparison. B1 is trained on the original random walk and its left-right mirrored copy. B2 uses the symmetry training of~\cite{mittal2024symmetry}, which combines a symmetry loss with symmetry data augmentation. The symmetry loss is added to the PPO loss, with weight $0.5$ in our case, and penalizes the difference between the action for a mirrored observation and the mirrored action. With data augmentation, every PPO batch is extended with its mirrored copy, where the observations and actions are mirrored, so the policy is also updated on the mirrored samples. B2 is trained on the original random walk only. B3 combines both: it uses the symmetry training of B2 and is trained on the original and the mirrored random walk as B1. We evaluate all variants on the plain random walk and on its left-right mirror.

\paragraph{Simulation.}
On the plain motion, all variants succeed in every environment and their MJPE is between $3.06^\circ$ and $3.17^\circ$ (Table~\ref{tab:sim_sym}). On the mirrored motion, B0 fails in almost half of the environments (success $54\%$), while B1, B2, and B3 reach $100\%$ with the same error as on the plain motion. Either the mirrored training data or the symmetry training is enough to avoid the failure of B0 on the mirrored motion.

\paragraph{Real robot.}
B1, B2, and B3 complete all three trials on both the plain and the mirrored random walk (Table~\ref{tab:real_sym}). B0 completes none of the three trials on the mirrored random walk, as expected from its low success in simulation. The tracking differences between B1, B2, and B3 are small (MJPE $3.43^\circ$ to $3.56^\circ$ on average), with B2 the lowest. These results agree with the simulation.

\subsection{Ablation: Smoothed Motions}
\label{sec:res_smooth}

\begin{table}[t]
\centering
\scriptsize
\setlength{\tabcolsep}{2.5pt}
\caption{Simulation ablation, smoothed training motion. Both policies are trained on the random walk only, with the plain (P) or the smoothed (S) base trajectory, and evaluated on the random walk and on the held-out backward, side, and slow walking of the same variant. Metrics are computed on the successful environments only. Bold: best value.}
\label{tab:sim_smooth}
\begin{tabular}{llccccc}
\toprule
Motion & Method & Succ.$\uparrow$ & MJPE$\downarrow$ & MJVE$\downarrow$ & MMD$\downarrow$ & COV$\uparrow$ \\
\midrule
\multirow{2}{*}{Random walk} & P$\rightarrow$P & \textbf{100\%} & \textbf{3.07} & \textbf{0.34} & \textbf{0.10} & \textbf{0.47} \\
 & S$\rightarrow$S & \textbf{100\%} & 3.46 & 0.37 & 0.12 & 0.44 \\
\midrule
\multirow{2}{*}{Backward} & P$\rightarrow$P & \textbf{100\%} & \textbf{4.05} & \textbf{0.41} & \textbf{0.14} & \textbf{0.34} \\
 & S$\rightarrow$S & 99\% & 4.71 & 0.49 & 0.16 & 0.32 \\
\midrule
\multirow{2}{*}{Side} & P$\rightarrow$P & \textbf{100\%} & \textbf{3.34} & \textbf{0.27} & \textbf{0.17} & \textbf{0.39} \\
 & S$\rightarrow$S & \textbf{100\%} & 3.99 & 0.33 & 0.19 & 0.38 \\
\midrule
\multirow{2}{*}{Slow} & P$\rightarrow$P & \textbf{98\%} & \textbf{6.20} & \textbf{0.43} & \textbf{0.35} & \textbf{0.25} \\
 & S$\rightarrow$S & \textbf{98\%} & 6.31 & 0.46 & 0.38 & 0.23 \\
\midrule
\multirow{2}{*}{Avg.} & P$\rightarrow$P & \textbf{100\%} & \textbf{4.16} & \textbf{0.36} & \textbf{0.19} & \textbf{0.36} \\
 & S$\rightarrow$S & 99\% & 4.62 & 0.41 & 0.21 & 0.34 \\
\bottomrule
\end{tabular}
\end{table}

\begin{table}[t]
\centering
\scriptsize
\setlength{\tabcolsep}{2.5pt}
\caption{Real-robot ablation, smoothed training motion. Both policies are trained on the random walk only, with the plain (P) or the smoothed (S) base trajectory, and evaluated on the random walk and on the held-out backward, side, and slow walking of the same variant. Metrics are computed on the successful trials only (mean $\pm$ std). Bold: best value.}
\label{tab:real_smooth}
\begin{tabular}{llccccc}
\toprule
Motion & Method & Succ.$\uparrow$ & MJPE$\downarrow$ & MJVE$\downarrow$ & MMD$\downarrow$ & COV$\uparrow$ \\
\midrule
\multirow{2}{*}{Random walk} & P$\rightarrow$P & \textbf{3/3} & \textbf{3.68}{\tiny$\pm$0.04} & \textbf{0.41}{\tiny$\pm$0.00} & \textbf{0.13}{\tiny$\pm$0.00} & 0.42{\tiny$\pm$0.00} \\
 & S$\rightarrow$S & \textbf{3/3} & 3.90{\tiny$\pm$0.01} & 0.46{\tiny$\pm$0.00} & \textbf{0.13}{\tiny$\pm$0.00} & \textbf{0.45}{\tiny$\pm$0.01} \\
\midrule
\multirow{2}{*}{Backward} & P$\rightarrow$P & \textbf{3/3} & \textbf{4.09}{\tiny$\pm$0.02} & \textbf{0.40}{\tiny$\pm$0.01} & \textbf{0.12}{\tiny$\pm$0.00} & \textbf{0.40}{\tiny$\pm$0.01} \\
 & S$\rightarrow$S & \textbf{3/3} & 4.74{\tiny$\pm$0.04} & 0.45{\tiny$\pm$0.01} & 0.14{\tiny$\pm$0.01} & 0.36{\tiny$\pm$0.00} \\
\midrule
\multirow{2}{*}{Side} & P$\rightarrow$P & \textbf{3/3} & \textbf{3.29}{\tiny$\pm$0.02} & \textbf{0.23}{\tiny$\pm$0.00} & \textbf{0.15}{\tiny$\pm$0.00} & \textbf{0.43}{\tiny$\pm$0.01} \\
 & S$\rightarrow$S & \textbf{3/3} & 3.88{\tiny$\pm$0.02} & 0.30{\tiny$\pm$0.01} & 0.16{\tiny$\pm$0.00} & 0.41{\tiny$\pm$0.01} \\
\midrule
\multirow{2}{*}{Slow} & P$\rightarrow$P & \textbf{3/3} & 5.96{\tiny$\pm$0.14} & 0.39{\tiny$\pm$0.02} & 0.37{\tiny$\pm$0.03} & \textbf{0.34}{\tiny$\pm$0.01} \\
 & S$\rightarrow$S & \textbf{3/3} & \textbf{5.65}{\tiny$\pm$0.12} & \textbf{0.34}{\tiny$\pm$0.01} & \textbf{0.29}{\tiny$\pm$0.02} & \textbf{0.34}{\tiny$\pm$0.00} \\
\midrule
\multirow{2}{*}{Avg.} & P$\rightarrow$P & \textbf{100\%} & \textbf{4.26} & \textbf{0.36} & 0.20 & \textbf{0.39} \\
 & S$\rightarrow$S & \textbf{100\%} & 4.54 & 0.39 & \textbf{0.18} & \textbf{0.39} \\
\bottomrule
\end{tabular}
\end{table}

The base velocity of the recorded motions oscillates with the gait, while a gamepad command is smooth. We therefore also train our policy on the random walk with a smoothed base trajectory. As for the plain policy, the random walk is the only training motion, and backward, side, and slow walking are held out. We smooth the base position and the frame-to-frame base rotation of each clip with a Savitzky-Golay filter \cite{savitzky1964smoothing} (window of 100 frames), and keep the joint trajectories unchanged. The joint errors are therefore computed against the same targets. Smoothing changes the velocity command and also the reference base position and orientation. Each policy is evaluated on the four motions with the base trajectory it was trained with: plain motions for the plain policy (P$\rightarrow$P) and smoothed motions for the smoothed policy (S$\rightarrow$S).

\paragraph{Simulation.}
Success remains high for both policies (Table~\ref{tab:sim_smooth}; $100\%$ for the plain and $99\%$ for the smoothed policy on average). With smoothed training, MJPE rises by $11\%$ ($4.16^\circ$ to $4.62^\circ$), MJVE rises by $14\%$, and the style changes slightly (MMD $0.19$ to $0.21$, COV $0.36$ to $0.34$).

\paragraph{Real robot.}
Both the plain and the smoothed policy complete all three trials on the random walk and on the three held-out motions (Table~\ref{tab:real_smooth}). As in simulation, the smoothed policy tracks slightly worse on average (MJPE $4.54^\circ$ vs.\ $4.26^\circ$) with a similar style (COV $0.39$ for both, MMD $0.18$ vs.\ $0.20$). On slow walking, the smoothed policy has a lower MJPE ($5.65^\circ$ vs.\ $5.96^\circ$), MJVE, and MMD, and a similar COV. Training on smoothed motions therefore costs little tracking accuracy and gives a policy that can follow smooth, gamepad-like velocity commands.

\section{Conclusion}
\label{sec:conclusion}

We presented a velocity-conditioned whole-body tracking policy that is trained directly with RL and an asymmetric actor-critic. The critic and the tracking reward use the full reference motion, while the actor observes only proprioception, the anchor orientation error, and a planar velocity command. On the motions it was trained on, our policy stays close to whole-body tracking, with a mean joint position error about $8\%$ higher in simulation. When trained on the random walk and commanded with velocity profiles from held-out motions, it is the only tracking-based method that keeps working on all of them, in simulation and on the Unitree G1, while whole-body tracking and command-conditioned distillation fail on slow walking. Compared with AMP, our policy has a higher style coverage and runs on the real robot without a safety crane. The input analysis shows that most velocity commands of the held-out motions are close to the training data, while far fewer of the reference joint frames are. \textcolor{black}{In the ablations, mirrored training data or symmetry training, alone or combined, avoids the failure of the baseline policy on mirrored motions, and training on a smoothed base trajectory costs little tracking accuracy, in simulation and on the real robot.}

\bibliographystyle{IEEEtran}
\bibliography{bibliography}

\end{document}